\documentclass{article}
\usepackage{spconf,graphicx}
\usepackage{amsmath,bm} 
\usepackage{cleveref}
\crefname{figure}{Fig.}{Figures}
\Crefname{figure}{Figure}{Figures}
\crefname{equation}{}{}
\crefname{table}{Table}{Tables}
\creflabelformat{equation}{#2\textup{#1}#3}
\usepackage{tabularx,ragged2e,booktabs,caption}
\usepackage{multicol,tabularx,capt-of}
\usepackage{hhline}
\usepackage{multirow}
\usepackage{array}
\usepackage{xcolor} 
\usepackage{subcaption}
\usepackage{amsfonts}
\usepackage{slashbox}
\usepackage{enumitem}
\usepackage[T1]{fontenc}
\usepackage[utf8]{inputenc}

\title{DEEP IMAGE SEGMENTATION VIA DISCRIMINANT FEATURE LEARNING}
\name{Adam Dawid Sztamborski$^{1,2,\dagger}$, Raül~Pérez-Gonzalo$^{1,\dagger}$, and Antonio~Agudo$^{1}$\thanks{Supported by GRAVATAR project (PID2023-151184OB-I00), which has been funded by MCIU/AEI/10.13039/501100011033 and ERDF, EU. \newline \indent $^{\dagger}$These authors contributed equally.}}
\address{$^{1}$Institut de Robòtica i Informàtica Industrial, CSIC-UPC, Barcelona, Spain\\$^{2}$Politechnika Łódzka, Łódź, Poland}
\begin{document}
%\ninept
%
\maketitle
\begin{abstract}
Accurate image segmentation remains challenging, particularly in generating sharp, confident boundaries. While modern architectures have advanced the field, many of them still rely on standard loss functions like Cross-Entropy and Dice, which often neglect the discriminative structure of learned features, leading to inaccurate boundaries. This work introduces Deep Discriminant Analysis (DDA), a differentiable, architecture-agnostic loss function that embeds classical discriminant principles for network training. DDA explicitly maximizes between-class variance while minimizing within-class one,  promoting compact and separable feature distributions without increasing inference cost. Evaluations on the DIS5K benchmark demonstrate that DDA consistently improves segmentation accuracy, boundary sharpness, and model confidence across various architectures. Our results show that integrating discriminant analysis offers a simple, effective path for building more robust segmentation models.
\end{abstract}
 %Boundary Refinement, Confidence Optimization

\section{Introduction}
Image segmentation represents a core image processing problem with applications across medical imaging, green energy and autonomous systems~\cite{dsa-ilora,hpl}. Recent progress has focused on increasing architecture complexity rather than revisiting the optimization objective. Hence, these methods still rely on standard loss objectives such as Binary Cross-Entropy (BCE)~\cite{bce_mao2023cross} and Dice~\cite{dice_loss_milletari2016v}, which just optimize pixel-wise similarity, resulting in blurred or uncertain boundaries. Advanced losses such as focal, Jaccard, and hybrid losses~\cite{bunet,terven2025comprehensive_survey_on_losses} mitigate class imbalance or promote boundary quality, yet they overlook the discriminative structure of feature representations.

Classical statistical methods such as Linear Discriminant Analysis (LDA) explicitly maximize the ratio of between-class to within-class variance, producing compact and well-separated class distributions. Integrating such discriminant principles into deep networks offers a principled way to enhance feature compactness, but remains largely unexplored for dense prediction problems such as segmentation.

%\section{RELATED WORK} 
%\subsection{Related Work}
\noindent \textbf{Image Segmentation.} U-Net~\cite{unet} popularized encoder–decoder structures for end-to-end pixel-level prediction using skip connections, forming the backbone of many segmentation architectures~\cite{bunet,image_segmentation_fusion}. Numerous U-Net variants~\cite{azad2023daeattention_segmentation,unet_with_attention} have further strengthened feature localization, multi-scale fusion, and boundary refinement through attention mechanisms, hierarchical guidance, and context aggregation~\cite{att_unet,r2unet}. Transformer-based designs extend this progress by modeling long-range dependencies~\cite{birefnet,mask2former}, with notable examples integrating attention with spatial hierarchies~\cite{chen97transunet} or complementary mechanisms to enhance semantic relevance and computational efficiency~\cite{u2net,dis5k,udun}. Foundation models such as SAM~\cite{sam} and SAM2~\cite{sam2} further broaden this landscape toward task-agnostic segmentation. 

%\subsection{Discriminant Analysis in Deep Learning}
\noindent \textbf{Learning-based Discriminant Analysis.}
Distinct methods integrate discriminant analysis principles into neural networks for classification, such as neural Fisher discriminant analysis~\cite{dda-related-papers_neural_fisher}, least-squares Fisher extension~\cite{dda-related-papers_lsfda}, and Deep LDA losses~\cite{dlda_loss, dlda_loss1, dalf}. These models improve class compactness and inter-class separability in high-dimensional embeddings; however, an explicit discriminant formulation embedded directly into the loss remains unexplored. 

\noindent \textbf{Deep Discriminant Analysis.} Addressing this gap, this paper proposes the Deep Discriminant Analysis (DDA) loss, a differentiable objective that embeds discriminant analysis principles into the network optimization, which unifies discriminant learning with end-to-end segmentation optimization to improve global coherence and decision boundary consistency. 

The main contributions of this work are:
\begin{itemize}[noitemsep, nolistsep]
\item A novel discriminant-based loss for deep networks, inspired by Fisher discriminant analysis~\cite{lda}.
\item Architecture-agnostic and lightweight formulation that adds no parameters, leaving inference cost unchanged.
\item Comprehensive empirical validation across multiple architectures and the DIS5K benchmark~\cite{dis5k}.
\item Better feature separability and border confidence, leading to sharper and more reliable segmentation results.
\end{itemize}
%By bridging classical discriminant analysis with modern deep learning, DDA offers a simple, architecture-agnostic step toward more interpretable and robust segmentation models.

%Extensive experiments across multiple encoder–decoder architectures were conducted, comparing against recent state-of-the-art models and standard baselines. DDA consistently demonstrates performance gains in classical segmentation metrics, confirming that enhancing feature discriminability can yield measurable segmentation improvements. 

%To sum up, DDA combines classical discriminant analysis with modern deep learning to offer a simple and interpretable approach that strengthens robustness and discriminability of segmentation models. Extensive experiments across multiple architectures were conducted, demonstrating DDA's performance gains and confirming that enhancing feature discriminability can yield measurable segmentation improvements. 

DDA combines classical discriminant analysis with deep learning to offer an elegant, interpretable approach to improve the robustness and discriminability of segmentation models. %Extensive experiments across multiple architectures were conducted, demonstrating DDA's performance gains and confirming that enhancing feature discriminability can yield measurable segmentation improvements. 

%====================================================
\section{Deep Discriminant Analysis}

% Pixel-wise losses such as BCE~\cite{bce_mao2023cross} and Dice~\cite{dice_loss_milletari2016v} treat pixels independently and do not constrain the global structure of feature distributions learned by the network. As a result, foreground and background activations may overlap, producing uncertain or blurred boundaries. To address this, we propose a differentiable DDA loss function derived from classical Fisher–style discriminant analysis. We first review LDA's statistical foundations, then we discuss separability measures based on scatter matrices that can be used in DDA formulation; and finally, we present the segmentation variant relevant for pixel-wise classification. %The aim of this section is to build a consistent mathematical bridge between classical discriminant theory and modern end-to-end learning. %naturally generalize to a deep-learning setting
 
Pixel-wise losses such as BCE~\cite{bce_mao2023cross} and Dice~\cite{dice_loss_milletari2016v} treat pixels independently and do not constrain the global structure of feature distributions learned by the network. It can lead to overlapping foreground and background activations and uncertain or blurred boundaries. To address this, we propose a differentiable DDA loss function derived from classical Fisher–style discriminant analysis. We first review LDA's statistical foundations, then discuss scatter-based separability measures for the DDA formulation, and finally, present the segmentation variant relevant for pixel-wise classification.

\subsection{Discriminant Analysis} 
Discriminant analysis offers a multivariate statistical framework for distinguishing predefined classes by identifying optimal discriminative hyperplanes or projections~\cite{gaussian-bayes-LDA}. While the Bayes classifier yields the minimum theoretical error rate, it requires known data distributions~\cite{book:fukunaga}. Linear Discriminant Analysis (LDA) approximates this by assuming normally distributed classes with equal covariances, identifying a projection that maximizes class separability. As such, LDA both performs classification and offers dimensionality reduction by linearly projecting data onto a discriminative subspace.

\noindent \textbf{The Fisher criterion.} Fisher discriminant analysis seeks a linear projection $d$-dimensional vector $\mathbf{w} \in \mathbb{R}^d$ that maximizes the ratio of between-class to within-class scatter~\cite{lda}. Unlike LDA, it does not assume that all class covariances are equal, making it robust when this assumption is violated~\cite{book:fukunaga}. 

Consider a classification problem with $n$ samples and $L$ classes, where class $C_k$ contains $n_k$ samples. Let $\mathbf{x}_i^{(k)} \in \mathbb{R}^d$ denote the $i$-th sample of class $C_k$. Given the class $C_k$, the mean vector $\boldsymbol{\mu}_k \in \mathbb{R}^d$ and covariance matrix $\mathbf{\Sigma}_k \in \mathbb{R}^{d\times d}$ are:
\vspace{-0.6cm}
\begin{align}
% Fukunaga 4.19, 2.31
\boldsymbol{\mu}_k &= \frac{1}{n_k}\sum_{i=1}^{n_k} \mathbf{x}_i^{(k)}, \\
% Fukunaga 4.20, 2.45
\mathbf{\Sigma}_k &= \frac{1}{n_k - 1}\sum_{i=1}^{n_k}  \left(\mathbf{x}_i^{(k)} - \boldsymbol{\mu}_k\right)
\left(\mathbf{x}_i^{(k)} - \boldsymbol{\mu}_k\right)^{\top} .
\end{align}
\vspace{-0.4cm}

For any input vector $\mathbf{x} \in \mathbb{R}^d$, the one-dimensional projection is defined by the affine mapping $y = \mathbf{w}^{\top} \mathbf{x} + w_0$, where $w_0 \in \mathbb{R}$ is a scalar bias term. Under this transformation, the projected class mean $m_k$ and projected class variance $s_k^2$ are:
\vspace{-0.6cm}
\begin{align}
    m_k &= \mathbf{w}^{\top} \boldsymbol{\mu}_k + w_0,\;\;
    s_k^2 = \mathbf{w}^{\top} \mathbf{\Sigma}_k \mathbf{w} .
\end{align}%\vspace{-0.55cm}
\vspace{-0.6cm}

\begin{figure}[!t]
% Addressing Antonio's comments
% Figure1: it is used w^T X, but according to text, should be w^T x, i.e, x instead of X. Moreover, use bold(w) rather than w. 
% \Adam: X changed to x, w made bold. Moreover, x made bold as well (just in case I upload an image with w bolded only; lda_by_bold-w.png).

    \centering
    \framebox{
      \parbox{3.3in}{
        \centering
        \includegraphics[width=1.\linewidth]{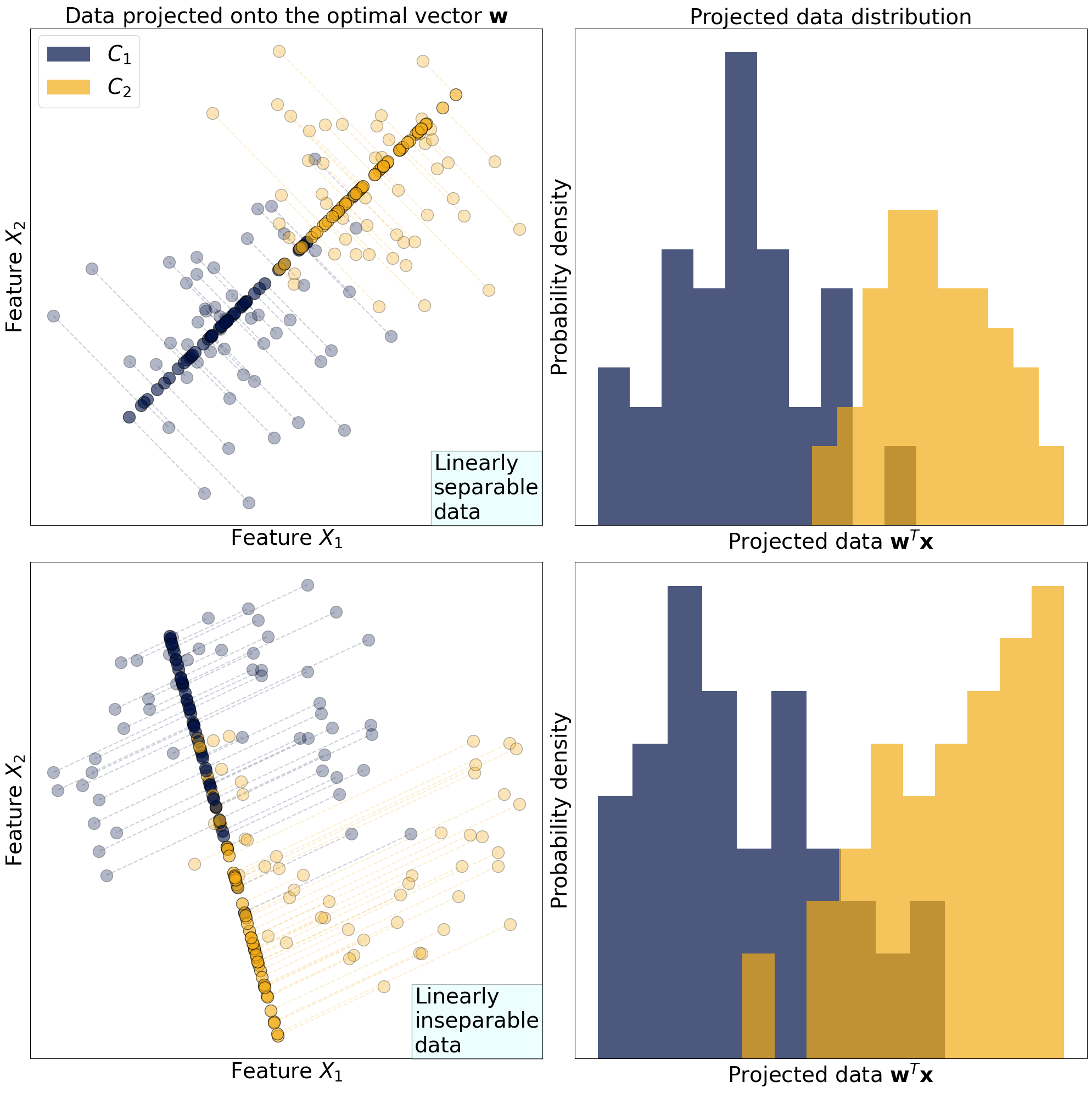}
      }
    }
    \vspace{-0.2cm}
    \caption{\textbf{Linear Discriminant Analysis application} on linearly separable (top) and inseparable (bottom) datasets.}    
    \label{fig:lda} 
    \vspace{-0.5cm}
\end{figure}

\Cref{fig:lda} illustrates optimal linear projections of linearly separable and inseparable classes for two-class data ($L=2$). In this case, the Fisher criterion $J$ maximizes class separability in the projected space and can be simply defined as:

\vspace{-0.2cm}
\begin{equation}
J(\mathbf{w}) = \frac{(m_1 - m_2)^2}{s_1^2 + s_2^2}.
\end{equation}
\vspace{-0.2cm}

Thus, the optimal projection of $\mathbf{w}$ is obtained by:
\vspace{-0.1cm}
\begin{align}
\mathbf{w}^\star = \arg\max_{\mathbf{w}} \left[J(\mathbf{w})\right].% = \underset{w}{argmax}\left[S_W^{-1}S_B\right]. % prince, eq. 6.1, pg 91/541 (77)%= \frac{S_B}{S_W}
\end{align}

However, only the optimal direction of $\mathbf{w}$ can be determined analytically, because the bias term $w_0$ cancels out in the difference $c=m_1 - m_2$ of projected class means:
\begin{equation}
c = \mathbf{w}^{\top} \boldsymbol{\mu}_1 + w_0 - \left(\mathbf{w}^{\top} \boldsymbol{\mu}_2 + w_0\right) = \mathbf{w}^{\top}(\boldsymbol{\mu}_1 - \boldsymbol{\mu}_2).
\end{equation}

To address this issue, Fukunaga~\cite{book:fukunaga} proposed a modified criterion that measures the between-class scatter (around zero) normalized by the within-class scatter and incorporates class priors $p_k \in \mathbb{R}$ that balance class separation as:
% scalars - lower case, so I change P_k to p_k
% TODO / TOCO / comment
% but how about the separability criterion J? It is a scalar value, but in the literature it is often denoted as J. A Similar case with L classes. Are they exceptions?
\vspace{-0.1cm}
\begin{align}
J(\mathbf{w}) &= \frac{p_1m_1^2 + p_2m_2^2}{p_1s_1^2+p_2s_2^2}. %= S_W^{-1}S_B \label{fukunaga_fisher_criterion}.
\end{align}
\vspace{-0.55cm}

\subsection{Making Discriminant Analysis Deep}

Real-world data such as images, speech, and biomedical signals often exhibit nonlinear structures that a single linear projection cannot fully capture. DDA overcomes this limitation by providing a bridge between LDA and the non-linear capabilities of deep-learning models. DDA embeds a Fisher-type separability criterion directly into the network training to encourage the statistical separation of learned features.

\noindent \textbf{General class separability.} We introduce a general class-separability criterion for an $L$-class classification problem based on the scatter statistics of projected vectors $\mathbf{y}_i^{(k)} \in \mathbb{R}^{L-1}$. Each $\mathbf{y}_i^{(k)} $ corresponds to the general projection of the input vector $\mathbf{x}_i^{(k)}\in \mathbb{R}^{d}$ from class $C_k$. The following scatter matrices characterize the separability of these projections:

1) The between-class scatter matrix $\mathbf{S}_{B}\in \mathbb{R}^{L-1\times L-1}$ measures how far the class mean vectors $\mathbf{m}_k\in \mathbb{R}^{L-1}$ lie from the global mixture mean $\mathbf{m}_0\in \mathbb{R}^{L-1}$. A larger $\mathbf{S}_{B}$ indicates better class separability. It can be estimated by:
\vspace{-2.0cm}
\begin{align}
% m_k as an unbiased estimate Fukunaga eq. 2.27,2.28,2.31,
% m_k as an unbiased estimate Fukunaga eq. 2.27,2.28,2.31,
{\mathbf{m}_k} &= \frac{1}{n_k}\sum_{i=1}^{n_k}\mathbf{y}_i^{(k)}, \;\;
{\mathbf{m}_0} = \sum_{k=1}^{L}\frac{n_k}{n}\mathbf{m}_k \label{m_0}, \\
\mathbf{S}_{B} &= \sum\limits_{k=1}^{L} \frac{n_k}{n}(\mathbf{m}_k-\mathbf{m}_0)(\mathbf{m}_k-\mathbf{m}_0)^{\top}.
\end{align}
\vspace{-0.4cm}

2) The within-class scatter matrix $\mathbf{S}_W \in \mathbb{R}^{(L-1)\times(L-1)}$ captures how tightly the samples of each class cluster are around their respective mean.
Smaller values indicate more compact, discriminable classes. It is defined as the prior-weighted sum of empirical class scatters $\mathbf{S}_k$:

\vspace{-0.6cm}
\begin{align}
% covariance matrix as an unbiased estimate Fukunaga eq. 10.3,
\mathbf{S}_k &= \frac{1}{n_k-1}\sum_{i=1}^{n_k}(\mathbf{y}_i^{(k)}-{\mathbf{m}_k})(\mathbf{y}_i^{(k)}-{\mathbf{m}_k})^{\top},
\label{dda-eq:within_class_scatter_matrix} \\
\mathbf{S}_{W} &= \sum\limits_{k=1}^{L}\frac{n_k}{n}\mathbf{S}_k  .%\\%\;\; 
\end{align}
\vspace{-0.4cm}

3) The covariance of all projected samples defines the mixture scatter matrix $\mathbf{S}_M\in \mathbb{R}^{L-1\times L-1}$, which naturally decomposes into within- and between-class terms:

\vspace{-0.6cm}
\begin{equation}
\mathbf{S}_M=\frac{1}{n-1}\sum_{i=1}^{n}(\mathbf{y}_i-{\mathbf{m}_0}) \notag(\mathbf{y}_i-{\mathbf{m}_0})^{\top}=\mathbf{S}_W+\mathbf{S}_B.
\end{equation}
\vspace{-0.4cm}

This desirable general separability criterion should take larger values when the within-class scatter is smaller and when the between-class scatter is larger. An efficient way to obtain such a separability measure is through the trace of the multiplied scatter matrices as $\textrm{tr}\{\mathbf{S}_{2}^{-1}\mathbf{S}_{1}\}$~\cite{book:fukunaga}. Among $\mathbf{S}_{B}$, $\mathbf{S}_{W}$, and $\mathbf{S}_{M}$, several valid choices of $\{\mathbf{S}_1,\mathbf{S}_2\}$ exist; in fact, they yield the same optimal projection for the linear case~\cite{book:fukunaga}. Hence, several DDA variants can be derived by choosing different pairs of scatter matrices, including:
\vspace{-0.2cm}
\begin{align}
(\mathbf{S}_{1}, \mathbf{S}_{2})&\in\{(\mathbf{S}_{B},\mathbf{S}_{W}), (\mathbf{S}_{W},\mathbf{S}_{M}), (\mathbf{S}_{B},\mathbf{S}_{M})\}.
\end{align}
\vspace{-0.65cm}

We adopt a particular combination of these scatter matrices to define our separability criterion $J(\mathbf{w})$: the multiplication of the inverse of the within-class scatter matrix $\mathbf{S}_{W}^{-1}$ with the between-class scatter matrix $\mathbf{S}_{B}$. Since all $L$ classes' mean vectors $\mathbf{m}_k$ are related via the mixture vector $\mathbf{m}_0$ (see Eq.~\eqref{m_0}), only $L-1$ of them are linearly independent. Consequently, both $\mathbf{S}_B$ and $\mathbf{S}_W^{-1}\mathbf{S}_B$ have rank $L-1$, yielding exactly $L-1$  nonzero eigenvalues. The separability criterion can therefore be expressed either via the trace or equivalently as the sum of these eigenvalues $\lambda_i$:

\vspace{-0.8cm}
\begin{align}  
J(\mathbf{w}) &= \textrm{tr}\{\mathbf{S}_{W}^{-1}\mathbf{S}_{B}\} = \sum\limits_{i=1}^{L-1} \lambda_i \label{dda-eq:separability_criterion_as_sum_of_eigens}.
\end{align}
\vspace{-0.4cm}

To encourage discriminative representations during deep learning model training, we define the DDA loss as the negative of this criterion:

\vspace{-0.5cm}
\begin{equation}
\mathcal{L}_{\mathrm{DDA}}(\mathbf{w}) = -\textrm{tr}\{\mathbf{S}_{W}^{-1}\mathbf{S}_{B}\} = -\sum\limits_{i=1}^{L-1} \lambda_i
\label{fukunaga_fisher_criterion_loss},
\end{equation}
where $\mathbf{S}_{W}^{-1}\mathbf{S}_{B}$ can be calculated from mini-batch samples for a non-linear system. In the binary case ($L=2$), this expression admits a closed-form reduction.

\noindent \textbf{Binary case.} In a binary image segmentation setting, the network outputs a single feature value per pixel (1-channel output), and each feature output is classified as a sample belonging to the foreground or the background. All scatter matrices, therefore, collapse to scalars, and the generalized eigenvalue problem reduces to a single eigenvalue. Using the two-class ANOVA decomposition~\cite{anova-decomp}, $ n_1 (m_1 - m_0)^2 + n_2 (m_2 - m_0)^2 = \frac{n_1 n_2}{n}(m_1 - m_2)^2,$ and the relation $\sum_{i=1}^{n_k}(y_i^{(k)}-m_k)^2=(n_k-1)s_k^2$, the binary DDA loss becomes:

\vspace{-0.5cm}
\begin{align} \label{dda_binary_loss}
\mathcal{L}_{\mathrm{DDA}}(\mathbf{w})
&= -\frac{n_1 n_2 (m_1 - m_2)^2}{n_1 s_1^2 + n_2 s_2^2}
= -\lambda_1.
\end{align}
\vspace{-0.3cm}

As a consequence, we can observe that $\mathcal{L}_{\mathrm{DDA}}$ fulfills that:
\begin{itemize} [noitemsep, nolistsep]
\item $\mathcal{L}_{\mathrm{DDA}}$ is differentiable with respect to $\mathbf{w}$.
\item The network itself becomes the nonlinear discriminant function when optimized with $\mathcal{L}_{\mathrm{DDA}}$.
\item In the binary case, $\mathcal{L}_{\mathrm{DDA}}$ avoids unnecessary complexity, keeping the loss lightweight and numerically stable. 
\item $\mathcal{L}_{\mathrm{DDA}}$ promotes separable feature distributions without altering the model architecture or inference cost.
\item $\mathcal{L}_{\mathrm{DDA}}$ offers an interpretable training signal, i.e., the numerator grows as class means separate, while the denominator shrinks as within-class variance tightens.
\end{itemize}

For pixel-wise segmentation, this loss explicitly encourages foreground and background activations to form well-separated distributions by leveraging pixel-value statistics across batches. This reduces the overlap of predicted mask logits, resulting in narrowed, well-separated class distributions. Empirically, this translates into sharper boundaries and improved robustness to noise and class imbalance.

%====================================================
\section{Experimental Evaluation}

% In this section, we discuss our experimental results from both quantitative and qualitative perspectives. In particular, our DDA loss was integrated into diverse backbone-free encoder–decoder segmentation architectures. These models employ distinct mechanisms, such as attention and residual blocks, for enhanced segmentation performance. For reference, we also evaluate zero-shot foundation models such as  SAM2~\cite{sam2} without task-specific fine-tuning. To isolate the impact of the loss function, we utilize the original implementation of each model. Following prior work~\cite{dis5k,udun}, we set $\beta^2=0.3$ for a generalized F-measure $F_\beta$.

In this section, we discuss our experimental results from both quantitative and qualitative perspectives. In particular, our DDA loss was integrated into diverse backbone-free encoder–decoder segmentation architectures. These models employ distinct mechanisms, such as attention and residual blocks, for enhanced segmentation performance. For reference, we also evaluate zero-shot foundation models such as  SAM2~\cite{sam2} without task-specific fine-tuning. To isolate the impact of the loss function, we utilize the original implementation of each model. 

\noindent \textbf{Metrics.} Following prior work~\cite{dis5k,udun}, we set $\beta^2=0.3$ for a generalized F-measure $F_\beta$. Boundary metrics~\cite{boundary_iou} are computed using boundary masks $B$, generated by subtracting an eroded version of the original mask $M$ from itself. The erosion ($\ominus$) is performed using a $3\times3$ structuring element $k$ for $d$ iterations, such that $B = M - (M \ominus_d k)$. To ensure consistent boundary thickness across varying resolutions, we adaptively scale the iterations as $d = \text{round}(0.02 \times \sqrt{H^2 + W^2})$, where $H$ and $W$ denote the image height and width, respectively. This approach preserves fine contour details while removing interior pixels, regardless of the input resolution.

\begin{table*}[t!] %!t
% \Adam -> With changes in Table 1 I want to cover:
% "[R3] Global separability vs boundary quality. We’d include boundary-specific metric in the final version, results of our experiments confirm that DDA strongly enhances boundary quality";
% "[R2, R3, R4] SAM comparison. As noted in Sec.III and Tab.1, SAM performance is used as a baseline reference for metric comparison, not for direct SOTA comparison. We used tight bounding boxes derived from ground-truth masks as prompts."
    \centering %{\backslashbox{Data}{Model}}
    \resizebox{0.85\textwidth}{!}{
    \small
    \setlength{\tabcolsep}{2pt}
    \renewcommand{\arraystretch}{.1}
    \begin{tabular}{l r|| c| c| c| c| c| c| c c}
        \toprule
        \multicolumn{2}{c||}{\backslashbox{Data}{Model}}   & U-Net~\cite{unet} & AttU-Net~\cite{att_unet} &R2U-Net~\cite{r2unet} & U$^2$-Net~\cite{u2net} & SAM*~\cite{sam} & SAM2.0*~\cite{sam2} & SAM2.1*~\cite{sam2} &\\
        \multicolumn{2}{r||}{Resolution} & {$512^2$} & {$512^2$} & {$224^2$} & {$1024^2$} & - & - & -\\
        \midrule \midrule

\multirow{4}{*}{DIS-TE1} % UNet, AttUnet, R2Unet, U2Net1024, Sam, Sam2, Sam2.1
& $IoU\space(bIoU) \uparrow$ & .424 (.407) & .454 (.429) & .397 (.364) & .637 (.619) & .671 (.617) & \textbf{.731} \textbf{(.669)} & .726 (.664)\\
& $F_1\space(bF_1) \uparrow$ & .565 (.553) & .590 (.571) & .529 (.499) & .736 (.728) & .753 (.713) & \textbf{.807} \textbf{(.765)} & .803 (.761)\\
& $F_\beta\space(bF_\beta) \uparrow$ & .643 (.607) & .670 (.636) & .640 (.588) & .796 (.772) & .796 (.740) & \textbf{.840} \textbf{(.777)} & .832 (.770)\\
& $AUC \uparrow$ & .921 & .925 & .779 & \textbf{.956} & .836 & .879 & .880\\
\midrule
\multirow{4}{*}{DIS-TE2} 
& $IoU\space(bIoU) \uparrow$ & .507 (.457) & .542 (.487) & .474 (.408) & \textbf{.695} \textbf{(.659)} & .629 (.567) & .648 (.583) & .659 (.589)\\
& $F_1\space(bF_1) \uparrow$ & .649 (.607) & .675 (.631) & .611 (.552) & \textbf{.790} \textbf{(.767)} & .722 (.680) & .739 (.696) & .749 (.702)\\
& $F_\beta\space(bF_\beta) \uparrow$ & .702 (.650) & .731 (.685) & .699 (.632) & \textbf{.840} \textbf{(.809)} & .779 (.723) & .783 (.718) & .789 (.720)\\
& $AUC \uparrow$ & .933 & .941 & .816 & \textbf{.956} & .817 & .836 & 0.847\\
\midrule
\multirow{4}{*}{DIS-TE3} 
& $IoU\space(bIoU) \uparrow$ & .539 (.479) & .580 (.517) & .498 (.423) & \textbf{.731} \textbf{(.686)} & .549 (.476) & .586 (.508) & .586 (.508)\\
& $F_1\space(bF_1) \uparrow$ & .682 (.631) & .713 (.665) & .639 (.572) & \textbf{.824} \textbf{(.795)} & .650 (.595) & .684 (.628) & .687 (.631)\\
& $F_\beta\space(bF_\beta) \uparrow$ & .721 (.670) & .756 (.706) & .708 (.639) & \textbf{.863} \textbf{(.832)} & .707 (.647) & .725 (.656) & .726 (.655)\\
& $AUC \uparrow$ & .935 & .948 & .819 & \textbf{.963} & .775 & .797 & .807\\
\midrule
\multirow{4}{*}{DIS-TE4} 
& $IoU\space(bIoU) \uparrow$ & .564 (.511) & .585 (.540) & .508 (.435) & \textbf{.724} \textbf{(.692)} & .505 (.397) & .528 (.429) & .536 (.429)\\
& $F_1\space(bF_1) \uparrow$ & .704 (.661) & .718 (.685) & .649 (.585) & \textbf{.817} \textbf{(.798)} & .618 (.526) & .642 (.559) & .652 (.562)\\
& $F_\beta\space(bF_\beta) \uparrow$ & .739 (.704) & .758 (.722) & .707 (.656) & \textbf{.862} \textbf{(.844)} & .637 (.578) & .653 (.596) & .660 (.597)\\
& $AUC \uparrow$ & .935 & .940 & .825 & \textbf{.960} & .770 & .787 & .799\\
\hline
\hline
\multirow{4}{*}{DIS-TE(1-4)} 
& $IoU\space(bIoU) \uparrow$ & .508 (.464) & .540 (.493) & .469 (.407) & \textbf{.697} \textbf{(.664)} & .589 (.514) & .623 (.547) & .627 (.548)\\
& $F_1\space(bF_1) \uparrow$ & .650 (.613) & .674 (.638) & .607 (.552) & \textbf{.792} \textbf{(.772)} & .686 (.629) & .718 (.662) & .723 (.664)\\
& $F_\beta\space(bF_\beta) \uparrow$ & .701 (.658) & .729 (.687) & .688 (.629) & \textbf{.840} \textbf{(.814)} & .730 (.672) & .750 (.687) & .752 (.685)\\
& $AUC \uparrow$ & .931 & .939 & .810 & \textbf{.958} & .800 & .824 & .833\\
\bottomrule
    \end{tabular}
    }
    \vspace{-0.3cm}
    \caption{\textbf{Quantitative evaluation on several models optimized using DDA loss.} Values in parentheses indicate the boundary-based counterpart of each metric~\cite{boundary_iou}. ($\uparrow$) indicates that higher values represent better performance. $^*$Zero-shot models were not fine-tuned on this dataset; tight bounding boxes derived from ground-truth masks were used as prompts.}
    \label{table:dda_sam_scores_table}
    \vspace{-0.5cm}
\end{table*}

\noindent \textbf{Dataset.} We evaluated our method on the public DIS5K benchmark~\cite{dis5k}, designed for the Dichotomous Image Segmentation (DIS) task focused on fine-grained, category-agnostic object extraction from natural images. The dataset comprises 5,470 high-resolution images (2k, 4k, and above) with detailed binary masks, spanning 225 categories across 22 groups, and is split into training (DIS-TR, 3,000 images), validation (DIS-VD, 470 images), and testing (DIS-TE, 2,000 images). DIS-TE is divided into four sets (from DIS-TE1 to DIS-TE4) based on their complexity. DIS5K~\cite{dis5k} provides a rigorous benchmark to examine whether our DDA loss facilitates network optimization and potentially improves two-label discrimination, generalizing across diverse, high-resolution natural scenes. Data augmentation, as outlined in the original paper, was also conducted. Ground-truth masks were binarized at a 0.5 threshold after downsizing during training, while full-resolution ones were used during validation and testing to report metrics.% Eqs.~\eqref{eq:prec-recall}-\eqref{m-eq:fbeta}.

\noindent \textbf{Implementation details.} Training was early-stopped based on validation convergence. AdamW optimizer~\cite{adamw} was used with learning rate $10^{-3}$, $\beta_1=0.9$, $\beta_2=0.999$, weight decay $0.01$, and $\epsilon=10^{-8}$, unless otherwise specified in the original publications. Experiments used an NVIDIA Tesla T4.

\begin{table*}[t!]
% \Adam -> TODO: Make it readable :D?

% With changes in Table 2 I want to address:
% "[R3] Global separability vs boundary quality. We’d include boundary-specific metric in the final version, results of our experiments confirm that DDA strongly enhances boundary quality";
% "[R1, R2, R3] Further comparison. We value this remark. Results for Dice loss will be added to Tab.2. Overall DIS-TE(1-4) results for brief insight (IoU/F1/FB): AttU-Net 512px: 0.505/0.645/0.702; R2U-Net 224px: 0.372/0.510/0.612; U2-Net 320px: 0.523/0.656/0.653. Compared to the data presented in our publication and below, DDA outperforms standard loss function."
% "[R1] Limited metrics in Tab.2. Tab.2 reported F-beta following DIS5K publications. We will include the following results that confirm the strength of DDA(IoU/F1)): U-Net 512px BCE: 0.459/0.602; U-Net 512px DDA: 0.508/0.650; AttU-Net 512px BCE: 0.523/0.660; AttU-Net 512px DDA: 0.540/0.674; R2U-Net 224px BCE: 0.474/0.615; R2U-Net 224px DDA: 0.469/0.607; U2-Net 320px BCE: 0.577/0.702; U2-Net 320px DDA: 0.577/0.704."

    \centering
    \small
    \resizebox{17.8cm}{!} {
    \begin{tabular}{l r || c c c c || c c c c || c c c c || c c c c  c c c c}
    \toprule
    \multicolumn{2}{c||}{\backslashbox{Data}{Model (Resolution)}}  & \multicolumn{3}{|c|}{U-Net~\cite{unet} (512$^2$)} & {${\Delta\%}$} & \multicolumn{3}{|c|}{AttU-Net~\cite{att_unet} (512$^2$)} & {${\Delta\%}$} & \multicolumn{3}{|c|}{R2U-Net~\cite{r2unet} (224$^2$)} & {${\Delta\%}$} & \multicolumn{3}{|c|}{U$^2$-Net~\cite{u2net} (320$^2$)}& {${\Delta\%}$} & {} \\
     \multicolumn{2}{r||}{Loss} & {$\mathcal{L}_{Dice}$} & {$\mathcal{L}_{BCE}$} & {${\mathcal{L}_{DDA}}$} & {} & {$\mathcal{L}_{Dice}$} & {$\mathcal{L}_{BCE}$} & ${\mathcal{L}_{DDA}}$ & {} & {$\mathcal{L}_{Dice}$} & {$\mathcal{L}_{BCE}$} & ${\mathcal{L}_{DDA}}$ & {} & {$\mathcal{L}_{Dice}$} & {$\mathcal{L}_{BCE}$} & ${\mathcal{L}_{DDA}}$ & {} \\
    \midrule
    \midrule

    \multirow{4}{*}{DIS-TE1} % Unet, AttUnet, R2AttUnet, U2Net 320px
    % {DIS-TE1} & { } % Dice, BCE, DDA 
    & $IoU\space(bIoU) \uparrow$ & .316 (.088) & .367 (.183) & .424 (.407) & \textbf{+15.3 (+122.4)} & .428 (.282) & .438 (.242) & .454 (.429) & \textbf{+3.7 (+52.1)} & .310 (.226) & .413 (.243) & .397 (.364) & -3.9 \textbf{(+49.8)} & .447 (.380) & .521 (.486) & .515 (.478) & -1.2 (-1.6)\\
    & $F_1\space(bF_1) \uparrow$ & .448 (.155) & .503 (.295) & .565 (.553) & \textbf{+12.4 (+87.5)} & .567 (.410) & .574 (.366) & .590 (.571) & \textbf{+2.8 (+39.3)} & .439 (.345) & .548 (.365) & .529 (.499) & -3.5 \textbf{(+36.7)} & .575 (.516) & .644 (.621) & .641 (.614) & -0.5 (-1.1)\\
    & $F_\beta\space(bF_\beta) \uparrow$ & .499 (.111) & .584 (.272) & .643 (.607) & \textbf{+10.1 (+123.2)} & .645 (.511) & .659 (.394) & .670 (.636) & \textbf{+1.6 (+24.5)} & .553 (.448) & .644 (.431) & .640 (.588) & -0.7 \textbf{(+31.2)} & .579 (.520) & .712 (.669) & .707 (.663) & -0.8 (-0.9)\\
    & $AUC \uparrow$ & .879 & .911 & .921 & \textbf{+1.1} & .862 & .930 & .925 & -0.5 & .722 & .909 & .779 & -14.3 & .893 & .944 & .939 & -0.5\\

    \midrule

    \multirow{4}{*}{DIS-TE2}
     % {DIS-TE2} & { }
    & $IoU\space(bIoU) \uparrow$ & .383 (.119) & .461 (.225) & .507 (.457) & \textbf{+10.0 (+103.1)} & .509 (.307) & .524 (.291) & .542 (.487) & \textbf{+3.4 (+58.6)} & .371 (.249) & .477 (.259) & .474 (.408) & -0.7 \textbf{(+57.5)} & .523 (.431) & .584 (.527) & .575 (.523) & -1.5 (-0.8)\\
    & $F_1\space(bF_1) \uparrow$ & .530 (.206) & .605 (.355) & .649 (.607) & \textbf{+7.3 (+71.0)} & .650 (.448) & .661 (.431) & .675 (.631) & \textbf{+2.1 (+40.8)} & .512 (.381) & .619 (.391) & .611 (.552) & -1.3 \textbf{(+41.2)} & .659 (.579) & .707 (.665) & .702 (.663) & -0.7 (-0.3)\\
     & $F_\beta\space(bF_\beta) \uparrow$ & .562 (.150) & .656 (.322) & .702 (.650) & \textbf{+7.0 (+101.9)} & .706 (.543) & .718 (.447) & .731 (.685) & \textbf{+1.9 (+26.2)} & .617 (.482) & .691 (.441) & .699 (.632) & \textbf{+1.1 (+31.1)} & .654 (.583) & .756 (.706) & .751 (.704) & -0.6 (-0.3)\\
     & $AUC \uparrow$ & .875 & .924 & .933 & \textbf{+1.0} & .880 & .941 & .941 & -0.1 & .750 & .920 & .816 & -11.3 & .895 & .948 & .945 & -0.3\\

    \midrule

    \multirow{4}{*}{DIS-TE3}
   % {DIS-TE3} & { }
   & $IoU\space(bIoU) \uparrow$ & .418 (.133) & .493 (.242) & .539 (.479) & \textbf{+9.4 (+97.9)} & .539 (.309) & .556 (.310) & .580 (.517) & \textbf{+4.2 (+66.8)} & .406 (.265) & .497 (.256) & .498 (.423) & \textbf{+0.2 (+59.6)} & .554 (.455) & .602 (.538) & .607 (.545) & \textbf{+0.8 (+1.3)}\\
    & $F_1\space(bF_1) \uparrow$ & .567 (.228) & .638 (.378) & .682 (.631) & \textbf{+6.8 (+66.9)} & .681 (.454) & .694 (.457) & .713 (.665) & \textbf{+2.7 (+45.5)} & .548 (.402) & .640 (.390) & .639 (.572) & -0.1 \textbf{(+42.3)} & .688 (.605) & .727 (.679) & .734 (.686) & \textbf{+0.9 (+1.0)}\\
    & $F_\beta\space(bF_\beta) \uparrow$ & .594 (.166) & .685 (.339) & .721 (.670) & \textbf{+5.1 (+97.6)} & .727 (.543) & .742 (.463) & .756 (.706) & \textbf{+1.9 (+30.0)} & .639 (.496) & .702 (.431) & .708 (.639) & \textbf{+1.0 (+28.8)} & .684 (.610) & .771 (.717) & .778 (.726) & \textbf{+1.0 (+1.3)}\\
    & $AUC \uparrow$ & .884 & .930 & .935 & \textbf{+0.6} & .884 & .947 & .948 & \textbf{+0.1} & .760 & .920 & .819 & -11.0 & .897 & .953 & .951 & -0.1 \\

    \midrule

    \multirow{4}{*}{DIS-TE4}
     % {DIS-TE4} & { }
    & $IoU\space(bIoU) \uparrow$ & .422 (.158) & .516 (.266) & .564 (.511) & \textbf{+9.2 (+92.1)} & .543 (.308) & .573 (.333) & .585 (.540) & \textbf{+2.0 (+62.2)} & .399 (.269) & .508 (.261) & .508 (.435) & -0.1 \textbf{(+61.7)} & .567 (.481) & .601 (.550) & .611 (.563) & \textbf{+1.7 (+2.4)}\\
    & $F_1\space(bF_1) \uparrow$ & .573 (.264) & .661 (.409) & .704 (.661) & \textbf{+6.6 (+61.6)} & .684 (.453) & .709 (.483) & .718 (.685) & \textbf{+1.3 (+41.8)} & .541 (.411) & .651 (.399) & .649 (.585) & -0.4 \textbf{(+42.3)} & .701 (.629) & .729 (.693) & .738 (.704) & \textbf{+1.3 (+1.6)}\\
    & $F_\beta\space(bF_\beta) \uparrow$ & .598 (.196) & .709 (.370) & .739 (.704) & \textbf{+4.2 (+90.3)} & .728 (.539) & .750 (.492) & .758 (.722) & \textbf{+1.1 (+34.0)} & .638 (.491) & .703 (.437) & .707 (.656) & \textbf{+0.6 (+33.6)} & .695 (.640) & .769 (.727) & .781 (.741) & \textbf{+1.5 (+1.9)}\\
    & $AUC \uparrow$ & .871 & .928 & .935 & \textbf{+0.7} & .871 & .943 & .940 & -0.3 & .747 & .915 & .825 & -9.9 & .895 & .946 & .945 & -0.1\\
 \hline
\hline
   \multirow{4}{*}{DIS-TE(1-4)}
   % {Overall DIS-TE(1-4)} & { }
    & $IoU\space(bIoU) \uparrow$ & .385 (.124) & .459 (.229) & .508 (.464) & \textbf{+10.7 (+102.6)} & .505 (.301) & .523 (.294) & .540 (.493) & \textbf{+3.3 (+63.8)} & .372 (.252) & .474 (.255) & .469 (.407) & -1.0 \textbf{(+59.6)} & .523 (.437) & .577 (.525) & .577 (.527) & 0.0 \textbf{(+0.4)} \\
    & $F_1\space(bF_1) \uparrow$ & .530 (.213) & .602 (.359) & .650 (.613) & \textbf{+8.0 (+70.8)} & .645 (.441) & .660 (.434) & .674 (.638) & \textbf{+2.2 (+44.7)} & .510 (.385) & .615 (.386) & .607 (.552) & -1.3 \textbf{(+43.0)} & .656 (.582) & .702 (.665) & .704 (.667) & \textbf{+0.3 (+0.3)} \\
    & $F_\beta\space(bF_\beta) \uparrow$ & .563 (.156) & .659 (.326) & .701 (.658) & \textbf{+6.5 (+101.8)} & .702 (.534) & .717 (.449) & .729 (.687) & \textbf{+1.6 (+28.7)} & .612 (.479) & .685 (.435) & .688 (.629) & \textbf{+0.5 (+31.3)} & .653 (.588) & .752 (.705) & .754 (.709) & \textbf{+0.3 (+0.6)}\\
    & $AUC \uparrow$ & .877 & .923 & .931 & \textbf{+0.8} & .874 & .940 & .939 & -0.2 & .745 & .916 & .810 & -11.6 & .895 &.948 & .945 & -0.2\\
  \midrule
    \end{tabular} }
    \vspace{-0.35cm}
    \caption{\textbf{Model comparison across different loss functions.} Values in parentheses indicate the boundary-based counterpart of each metric~\cite{boundary_iou}. ($\uparrow$) indicates that higher values represent better performance. Relative percentage difference (${\Delta}\%$) is calculated between DDA and the second-best loss. Model input resolutions follow the prior papers.}
    \label{table:losses_scores_table}
    \vspace{-0.5cm}
\end{table*} %boundary_f

%& $F_\beta \uparrow$

\begin{figure}[t!]
    \centering
   \resizebox{8.7cm}{!} { % why was it 8.4cm? :D
   \begin{tabular}{c}
        \includegraphics[clip, angle=0,width=1.0\linewidth]{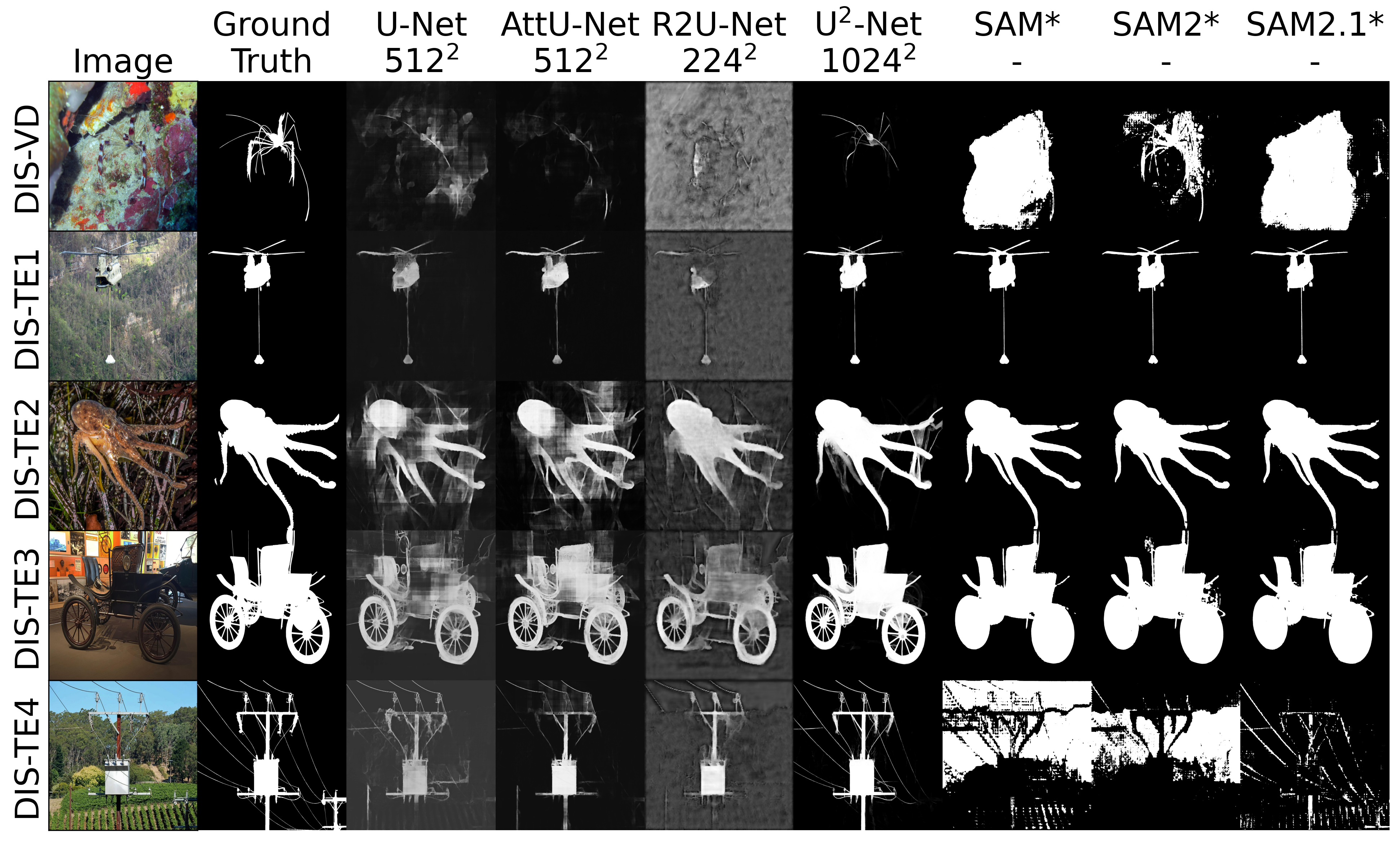} \vspace{-0.2cm} \\
        \includegraphics[clip, angle=0,width=1.0\linewidth]{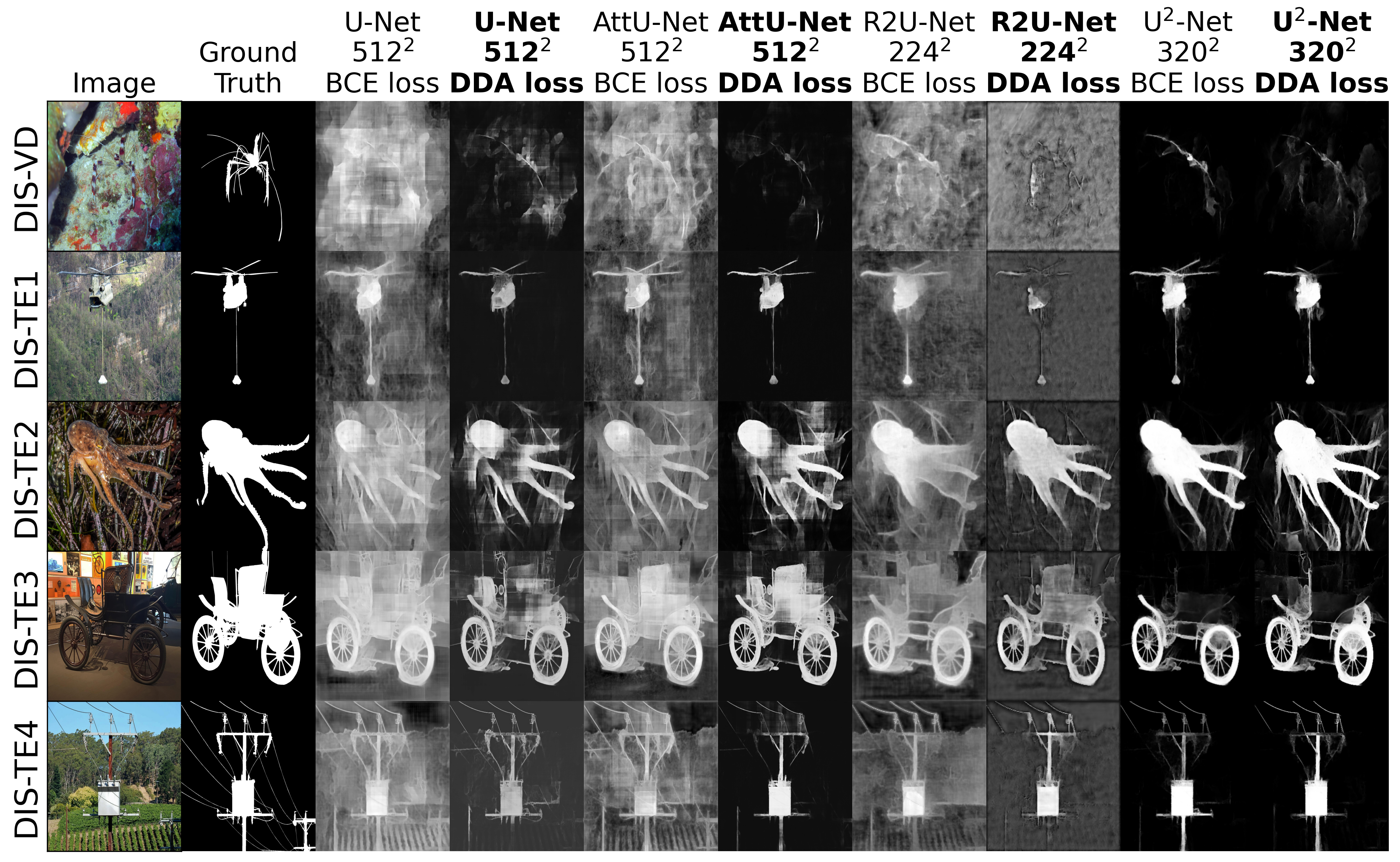}
    \end{tabular}
        }
    \vspace{-0.4cm}
    \caption{\textbf{Qualitative comparison between models} from Tables~\ref{table:dda_sam_scores_table} (top) and~\ref{table:losses_scores_table} (bottom). Best in color and with zooming.} \vspace{-0.4cm}
\label{fig:qualitative_comparison_between_architectures}
\end{figure}
    % to make the description consistent with figure 3 and liberate one line of text
    %\caption{\textbf{Qualitative comparison between models} from \cref{table:dda_sam_scores_table} optimized by DDA loss. Best viewed in color and with zooming.}

\noindent \textbf{Impact of DDA loss.} Our results are summarized in \cref{table:dda_sam_scores_table} and \cref{fig:qualitative_comparison_between_architectures}. As shown, zero-shot general-purpose SAM models~\cite{sam,sam2} are competitive on simpler case scenarios such as DIS-TE1, but degrade on more complex ones, due to the lack of task-specific optimization. However, the four models we trained with our DDA loss yielded maintained or improved scores, particularly in complex scenarios (DIS-TE2/3/4). 

In particular, U$^2$-Net architecture~\cite{u2net} provided the best solutions on average.  \Cref{fig:qualitative_comparison_between_architectures} depicts a representative failure case from every test set. Failure cases occur primarily when images exhibit very low contrast or highly cluttered backgrounds, where even more advanced models struggle (e.g., the image of well camouflaged shrimp from the DIS-VD set).
% TOCO : example from DIS-VD as failure case. Alternative (1): DIS-VD does not appear in tables and could be removed from figures as well - currently DIS-VD in figures serves only as an example of a failure case, a similar case appears in TE4 (I have just uploaded it) so it could solve it. Alternative (2): DIS-VD removed from figures and failure case displayed as separate Figure (there will be more space after cutting a DIS-VD rows from Figure 2 and 3), also uploaded a moment ago to ./ Figures / qualitative_comparison / failure_case_DIS_TE4_shrimp.png.

%To assess the generalizability of the proposed solution, we trained, using our DDA loss function, four distinct segmentation architectures on the DIS5K~\cite{dis5k} benchmark: U-Net~\cite{unet}, Attention U-Net (AttU-Net)~\cite{att_unet}, R2U-Net~\cite{r2unet}, and U$^2$-Net~\cite{u2net}. Their quantitative results are summarized in \cref{table:dda_sam_scores_table}, while visual comparisons appear in \cref{fig:qualitative_comparison_between_architectures}. 

\begin{figure}[t!]
% Addressing Antonio's comments
% Figure 4: Distributions of predicted mask pixel values for the vehicle image in Fig. 3. Within the figure, use: predicted pixel values. Title in the top part can be removed as this information is included in the caption. 
% \Adam: Done (histograms_.png. Just in case I upload histograms with Dice (histograms_dice_side_by_side.png, histograms_dice_stacked.png.
    \centering
  \resizebox{8.7cm}{!} {
        \includegraphics[clip, angle=0,width=1.0\linewidth]{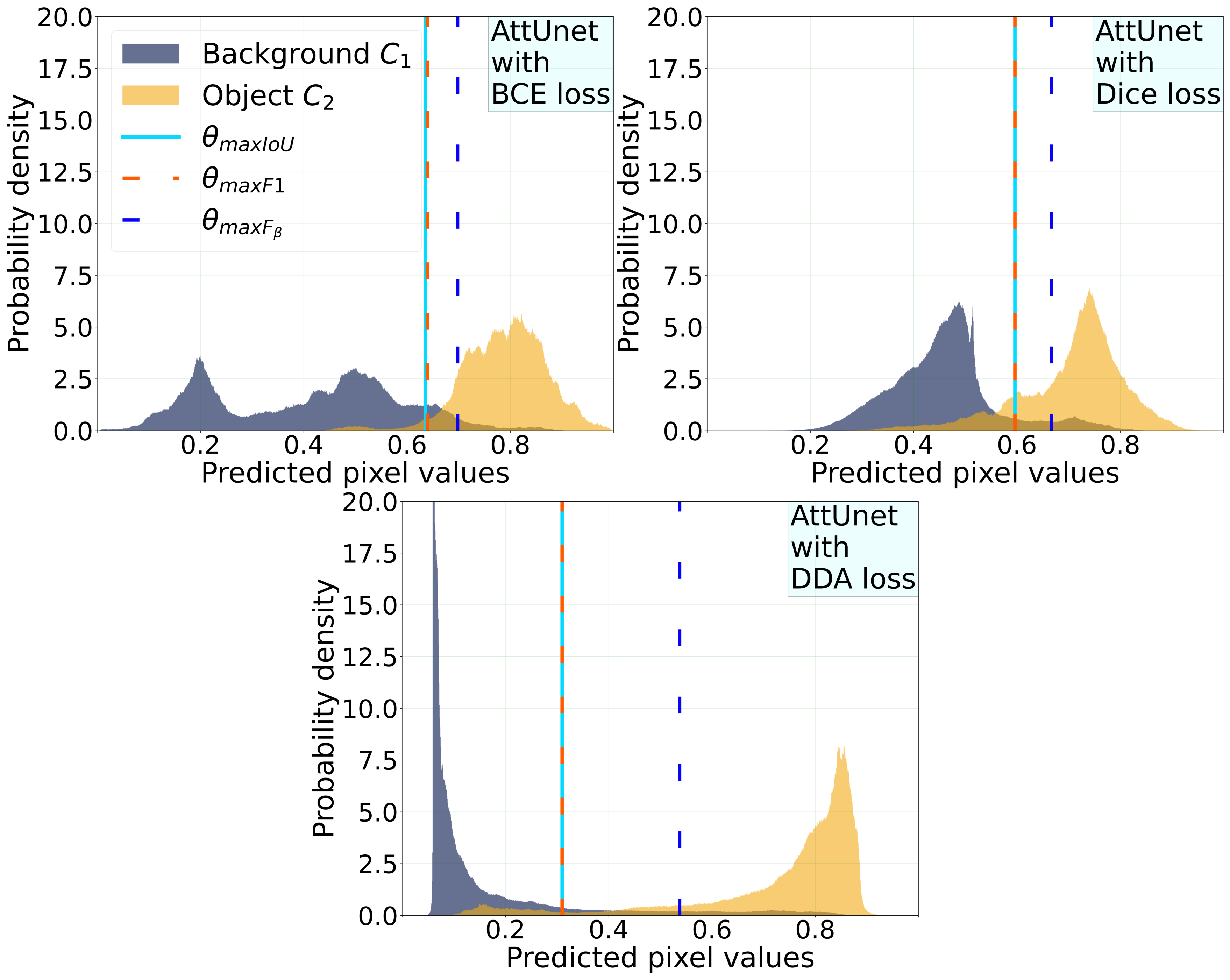} }
        \vspace{-0.5cm}
    \caption{\textbf{Distributions of predicted mask pixel values} of the vehicle image in \cref{fig:qualitative_comparison_between_architectures}-bottom. Vertical lines point the optimal thresholds $\theta$ maximizing given scores ($IoU$, $F_1$ or $F_{\beta}$).}
    \label{fig:histograms1}
    \vspace{-0.5cm}
\end{figure}

% For an additional reference point, we evaluated these architectures with the commonly used BCE and Dice losses. \Cref{table:losses_scores_table} compares their performance with DDA counterparts. In this comparison, U-Net~\cite{unet} and AttU-Net~\cite{att_unet} achieved the largest relative gains ($\mathbf{\Delta}\%$), while R2U-Net~\cite{r2unet} and U$^2$-Net~\cite{u2net} reported moderate improvements.  

For an additional reference point, we evaluated these architectures with BCE and Dice losses. \Cref{table:losses_scores_table} compares their performance with DDA counterparts. In this comparison, U-Net~\cite{unet} and AttU-Net~\cite{att_unet} achieved the largest relative gains ($\mathbf{\Delta}\%$), while R2U-Net~\cite{r2unet} and U$^2$-Net~\cite{u2net} reported moderate improvements. Notably, improvements in boundary metrics are often proportionally larger, suggesting that DDA improves not only foreground-background discrimination but also boundary consistency and contour localization. \Cref{fig:qualitative_comparison_between_architectures}-bottom reveals that DDA-trained models produce sharper, more accurate predictions with higher confidence.

Statistical insight in \cref{fig:histograms1} confirms alignment with DDA theoretical aspects; DDA approach demonstrates greater separation between classes (increased between-class variance) and narrower distributions (reduced within-class variance), resulting in less overlap. Optimal threshold lines for metrics are closer to the center, reflecting DDA’s emphasis on class separation around zero (after normalization, around $0.5$).

Finally, we compare the best combination in our previous experiments, U$^2$-Net~\cite{u2net} with our proposed DDA loss, with respect to recent state-of-the-art segmentation models: IS-Net~\cite{dis5k} and UDUN~\cite{udun}. In terms of $F_\beta$, U$^2$-Net~\cite{u2net} with DDA obtains the best performance in almost all datasets, and also on average; see results in \cref{table:compared_to_publications}.

% To sum up, replacing BCE or Dice with DDA in \cref{table:losses_scores_table} yields consistent improvements, suggesting that DDA effectively compensates for the limitations of simpler models and can enhance more complex ones. Whereas performance differences between architectures reflect mainly architectural capacity, as indicated by prior analyses~\cite{u2net,dis5k,udun}, we believe that this is due to our DDA loss, which represents a key factor in the final performance. 
To sum up, replacing BCE or Dice with DDA in \cref{table:losses_scores_table} yields consistent improvements, suggesting that DDA can compensate for the limitations of simpler models and enhance more complex ones. Whereas performance differences between architectures mainly reflect architectural capacity~\cite{u2net,dis5k,udun}, our results indicate that DDA is a key contributing factor to the final performance.

%%%%%%%%%%%%%%% COMPARISON TO PAPERS TABLE
\begin{table}[t!] %!t
    \centering
    \small
    \setlength{\tabcolsep}{1pt}
    \renewcommand{\arraystretch}{0.5}
    \begin{tabular}{l|| c c| c |c}
    \toprule
    {\backslashbox{Data}{Model}} &  \multicolumn{2}{c|}{U$^2$-Net~\cite{u2net}} & {IS-Net~\cite{dis5k}} & {UDUN~\cite{udun}}\\
     & { } & {+DDA} & { } & { }\\
     %& {Resolution} & {1024$^2$} & {1024$^2$} & {1024$^2$} & {1024$^2$}\\
    \midrule
    \midrule
     {DIS-VD}
     & .753 & \textbf{.853} & .791 & .823\\
    \midrule
    {DIS-TE1}
    & .701 & \textbf{.796} & .740 & .784\\
    \midrule
    {DIS-TE2}
    & .768 & \textbf{.840} & .799 & .829\\
    \midrule
    {DIS-TE3}
    & .813 & .863 & .830 & \textbf{.865}\\
    \midrule
    {DIS-TE4}
    & .800 & \textbf{.862} & .827 & .846\\
    \hline
    \hline
   {DIS-TE(1-4)}
    & .771 & \textbf{.840} & .799 & .831\\
  \midrule
    \end{tabular}
    \vspace{-0.35cm}
    \caption{\textbf{U$^2$-Net~\cite{u2net} with DDA compared to state-of-the-art models.} The table reports the $F_{\beta}$ score. Performance is evaluated at a resolution of $1024 \times 1024$.} 
    \vspace{-0.4cm}
    \label{table:compared_to_publications}
\end{table}

\section{Conclusion}
\vspace{-0.1cm}
In this work, we introduced deep discriminant analysis, a differentiable loss function that integrates classical Fisher discriminant principles into neural network optimization. By explicitly maximizing between-class variance and minimizing within-class variance, DDA enforces compact, separable feature representations without increasing model complexity or inference costs. Experiments on the DIS5K benchmark demonstrate that DDA is architecture-agnostic and consistently yields sharper boundaries and higher segmentation confidence. Notably, our DDA-optimized U$^2$-Net outperformed recent state-of-the-art methods, proving that embedding statistical discriminative criteria can significantly enhance modern deep-learning pipelines.

%\clearpage
\bibliographystyle{IEEEbib}
\bibliography{strings}

\end{document}